\documentclass[sigconf]{acmart}
\usepackage{xcolor}
\usepackage[mathscr]{eucal}
\usepackage{bm}

\AtBeginDocument{%
  \providecommand\BibTeX{{%
    \normalfont B\kern-0.5em{\scshape i\kern-0.25em b}\kern-0.8em\TeX}}}

\copyrightyear{2021} 
\acmYear{2021} 
\setcopyright{acmcopyright}\acmConference[DocEng '21]{ACM Symposium on Document Engineering 2021}{August 24--27, 2021}{Limerick, Ireland}
\acmBooktitle{ACM Symposium on Document Engineering 2021 (DocEng '21), August 24--27, 2021, Limerick, Ireland}
\acmPrice{15.00}
\acmDOI{10.1145/3469096.3474927}
\acmISBN{978-1-4503-8596-1/21/08}

\begin{document}

\title{COVID-19 Multidimensional Kaggle Literature Organization}


\author{Maksim E. Eren}
\authornote{Department of Computer Science and Electrical Engineering, UMBC}
\email{meren1@umbc.edu}
\affiliation{\country{USA}}

\author{Nick Solovyev}
\authornotemark[1]
\email{sonic1@umbc.edu}
\affiliation{\country{USA}}

\author{Chris Hamer}
\authornotemark[1]
\email{chamer1@umbc.edu}
\affiliation{\country{USA}}

\author{Renee McDonald}
\authornotemark[1]
\email{rp53139@umbc.edu}
\affiliation{\country{USA}}

\author{Boian S. Alexandrov}
\authornote{Theoretical Division, Los Alamos National Laboratory}
\email{boian@lanl.gov}
\affiliation{\country{USA}}

\author{Charles Nicholas}
\authornotemark[1]
\email{nicholas@umbc.edu}
\affiliation{\country{USA}}

\renewcommand{\shortauthors}{Eren et al.}

\begin{abstract}
The unprecedented outbreak of Severe Acute Respiratory Syndrome Coronavirus-2 (SARS-CoV-2), or COVID-19, continues to be a significant worldwide problem. As a result, a surge of new COVID-19 related research has followed suit. The growing number of publications requires document organization methods to identify relevant information. In this paper, we expand upon our previous work with clustering the CORD-19 dataset by applying multidimensional analysis methods. Tensor factorization is a powerful unsupervised learning method capable of discovering hidden patterns in a document corpus. We show that a higher-order representation of the corpus allows for the simultaneous grouping of similar articles, relevant journals, authors with similar research interests, and topic keywords. These groupings are identified within and among the latent components extracted via tensor decomposition. We further demonstrate the application of this method with a publicly available interactive visualization of the dataset.
\end{abstract}



\maketitle
\section{Introduction}
The world continues the fight against the Severe Acute Respiratory Syndrome Coronavirus-2 (SARS-CoV-2), or COVID-19. Now, more than a year removed from the start of the pandemic, COVID-19 research has not dwindled. As of July 2021, a Google Scholar search for "covid 19" shows 117,000 new publications in the last seven months.  The online statistical document analysis tool, \textit{Dimensions Database}, estimates over 534,000 new COVID-19 publications since 2020 \cite{Dimensions}. This overwhelming quantity of data makes the discovery of better document organization methods even more urgent. Document organization methods can assist with new research through information retrieval and sharing. 

The focus of this paper is organizing the \textit{COVID-19 Open Research Dataset} (CORD-19) \cite{Wang2020CORD19TC}\footnote{Dataset is available at \url{https://www.kaggle.com/allen-institute-for-ai/CORD-19-research-challenge}}. CORD-19 is a collection of over 400,000 scholarly articles about COVID-19 and related diseases. In our previous work on this dataset, we showed that investigation of the CORD-19 corpus can be simplified through clustering and dimensionality reduction using \textit{t-SNE}, \textit{PCA}, and \textit{k-means} \cite{eren2020}. The Kaggle notebook from our prior research has attracted great interest in the data science community\footnote{Kaggle notebook for the prior work is available at  \url{https://www.kaggle.com/maksimeren/covid-19-literature-clustering}}. In this paper, we continue to tackle the CORD-19 organization problem with a different approach by applying a multidimensional analysis method.


 Tensor decomposition is an unsupervised learning method capable of extracting multifaceted latent patterns from a dataset. Scientific papers are a type of data that is naturally multidimensional and can be represented as a tensor. Analyzing documents in a higher dimensional space allows for simultaneously finding correlations across each dimension. This approach provides for a more natural representation of the data in comparison to traditional matrix factorization methods. Specifically, we build a tensor with dimensions corresponding to author, title, journal, and keywords in the paper to characterize the documents in the CORD-19 dataset. 


In our work, we show that authors with similar research interests, relevant articles, and related journals can be grouped in and among the latent components through tensor factorization. At the same time, by representing the corpus vocabulary with keywords as a tensor dimension, we can identify the topic keywords for the papers. To the best of our knowledge, we are the first to use tensor analysis to organize the CORD-19 dataset. Finally, we present our results on a publicly available interactive visualization of the components we extracted via tensor decomposition\footnote{Interactive visualization is available at \url{https://maksimekin.github.io/CORD19-Tensor/}}.

\section{Relevant Work}
\label{sec:relevant_work}
Tensor and matrix decomposition and their application to text analysis is an area that has been widely studied. In this section we present a brief summary of related research.

An important and difficult problem for both matrix and
tensor decomposition
is determining the number of latent topics in a corpus. Vangara et al. factorize an ensemble of term frequency inverse document frequency (TF-IDF) matrices and apply k-means clustering to identify the rank of those matrices \cite{Vangara2020}, using a distributed software package \cite{pyDNMFk1, pyDNMFk3}. In their work, the number of latent topics is chosen to be the rank that returns the best combination of a high \textit{silhouette score} and a low \textit{relative reconstruction error} for the factorization. In comparison, we utilize the \textit{cosine similarity} score to identify the distinct topics over an ensemble of ranks rather than identifying a single optimal rank.

Larsen et al. introduce random sampling methods for faster computation of large tensors in \cite{larsen2020practical}. They apply their methods to analyse \textit{Reddit}\footnote{\textit{Reddit} is an online content sharing platform.} posts using a tensor with dimensions \textit{Subreddit} x \textit{User} x \textit{Word}, where the entries in the tensor are $log(1 + \text{word count})$. They observe that when the highest values in latent factors for \textit{Subreddit} dimension are plotted, similar subreddits\footnote{\textit{Subreddits} are topic specific discussion boards in \textit{Reddit}.} group together in the components. Similarly, we extract $n$ elements with the highest values in the latent factors to determine groupings, and use the log of word counts as tensor entries.

Tensor decomposition has been applied to medical data in previous studies including identification of chronic diseases \cite{Wang810556} and analysis of neurodevelopmental disorders \cite{hamdi2018}. Tensors have also been used to organize biomedical texts. Drakopoulos et al. built a tensor with the dimensions \textit{Term} x \textit{Keyword} x \textit{Document} which is a generalization of the \textit{term-document} matrix. They use TF-IDF values as tensor entries, and the clustering is done using k-means \cite{Drakopoulos2017}. In our work, we perform analysis over a four-dimensional tensor and find that components extracted via factorization can separate the documents, authors, and journals into groups and extract topic keywords.

Several document analysis methods applied to the CORD-19 dataset have been presented previously \cite{Grotheer2020}. Grotheer et al. demonstrated the use of hierarchical non-negative matrix factorization in \cite{Grotheer2020}. In their analysis, the corpus is represented as a \textit{Word} x \textit{Document} matrix. Given a number of topics, they decomposed the matrix into the \textit{dictionary} and \textit{coding} matrices. From there, the documents can be sorted into topic matrices using the \textit{coding matrix}. The process can be repeated with the "leaf" matrices codifying the corpus into a hierarchical tree \cite{Grotheer2020}. This approach, while yielding promising results on the CORD-19 dataset, is limited since the information is represented in two dimensional space. With higher dimensional analysis, we can observe the latent information regarding each dimension including the topic keywords which get extracted in an unsupervised manner. 

The remainder of the paper consists of a description of our tensor decomposition methods, the results of applying those methods to the CORD-19 dataset, and a concluding discussion regarding the effectiveness of our strategy.

\section{Methods}
Most of the pre-processing steps presented in Section \ref{subsec:pre_processing} improve upon the data cleaning methods from our prior work on the same dataset \cite{eren2020}. Then, Section \ref{subsec:tensor} describes the details of tensor construction and analysis of the latent factors.

\subsection{Data Pre-processing}
\label{subsec:pre_processing}
Before cleaning, the CORD-19 corpus consists of over 400,000 scholarly articles. We first drop the documents that lack a text body and/or are written in a language other than English. 
Next, the title and the journal name are shifted to lower-case, and numbers and special characters are removed. The duplicate papers (i.e. documents with the same title or documents with the same abstract) are dropped. After these steps, we are left with 128,359 unique articles published in 10,321 journals by 105,300 distinct first authors in the corpus. The stop-words are then removed from the articles and the text is tokenized using \text{SpaCy} biomedical 
\footnote{SciSpacy is used for text tokenization.} \cite{neumann2019} and English parsers \cite{spacy}. The word tokens are used to remove names, words with numerical values, and special characters. We then identify and remove DNA sequence patterns using the standard Python Regex function. Finally, we remove typos and nonsense words using the \textit{nonsense} Python library \cite{Hucka_2019, Hucka2018}. The SpaCy library was also used to perform lemmatization. After the text is processed, the documents in the dataset contain a total of 821,410 unique words.


\subsection{Tensor Factorization}
\label{subsec:tensor}

\begin{figure*}[t!]
\centering
\begin{minipage}[b]{.165\textwidth}
\includegraphics[width=\textwidth]{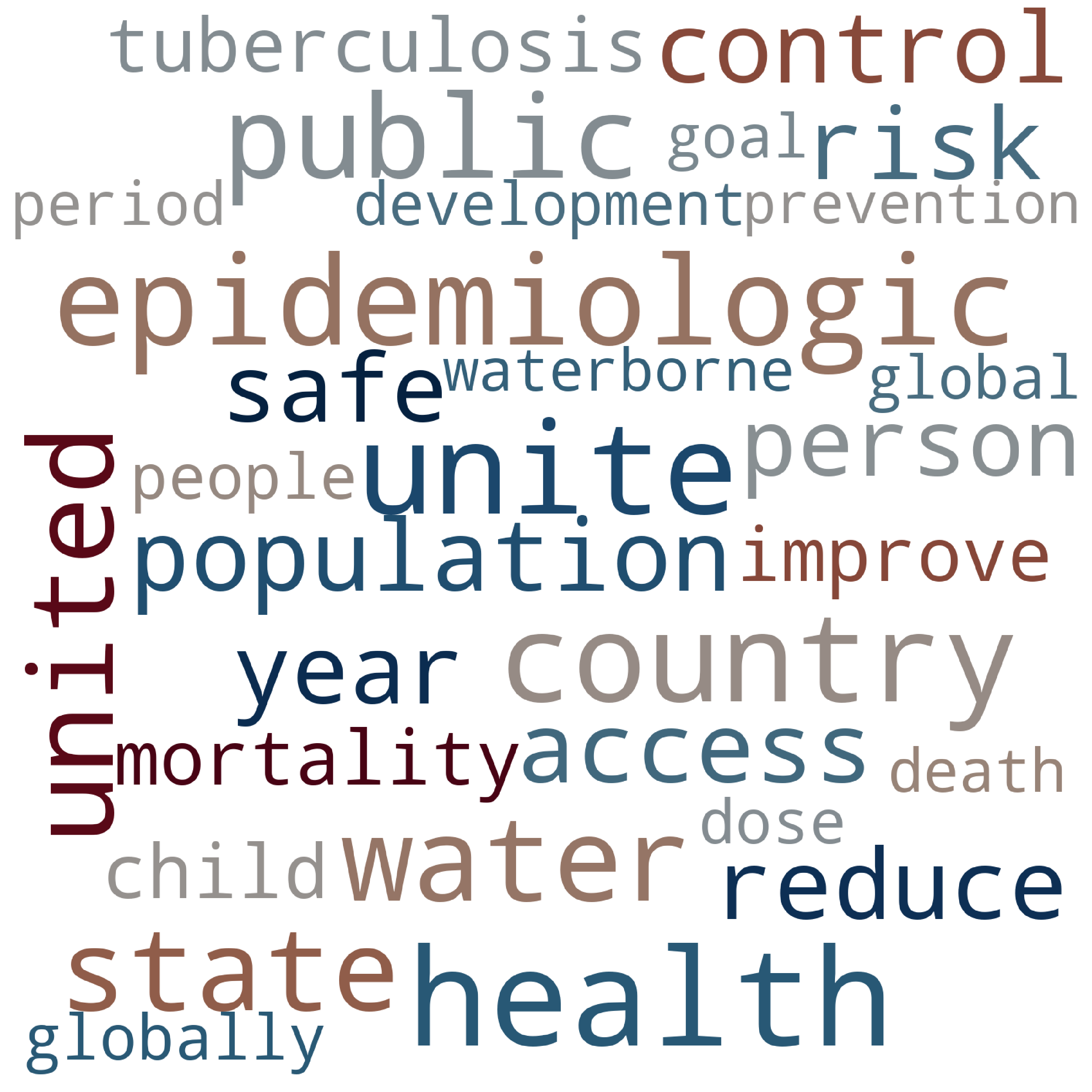}
\captionsetup{justification=centering} 
\vspace*{-6mm}
\caption{Component 7}\label{fig:c7_wc}
\end{minipage}\qquad
\begin{minipage}[b]{.165\textwidth}
\includegraphics[width=\textwidth]{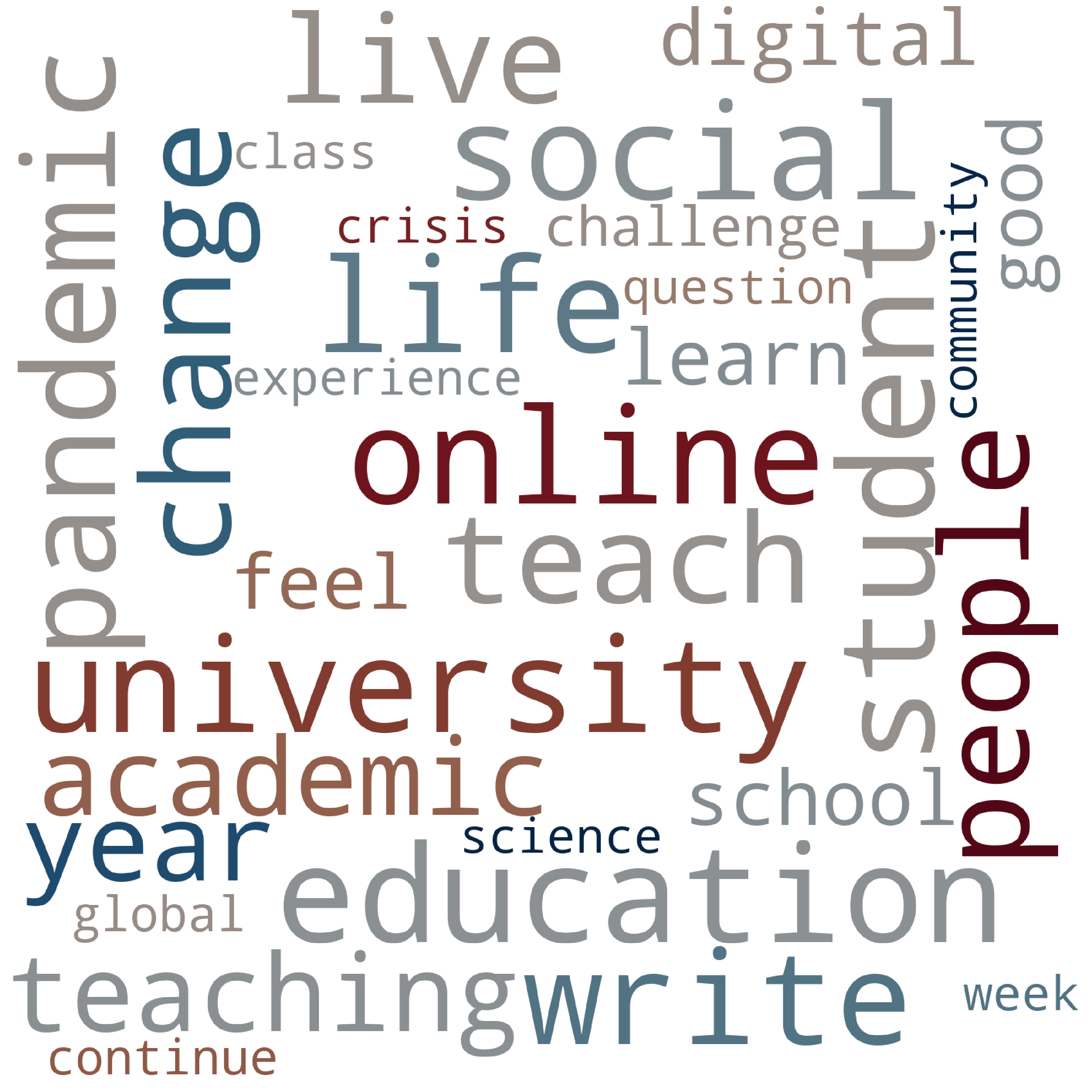}
\captionsetup{justification=centering}
\vspace*{-6mm}
\caption{\\Component 16}\label{fig:c16_wc}
\end{minipage}\qquad
\begin{minipage}[b]{.165\textwidth}
\includegraphics[width=\textwidth]{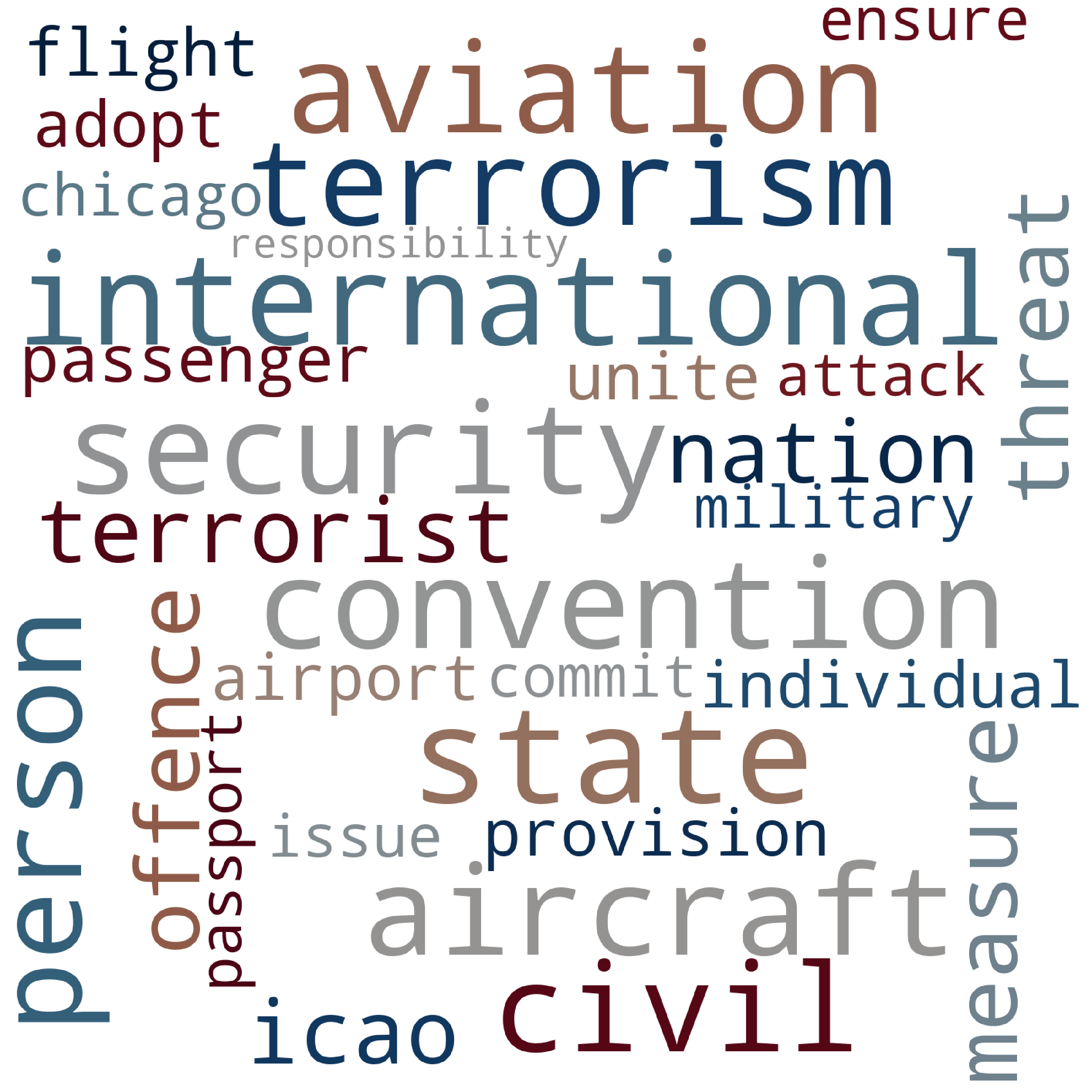}
\captionsetup{justification=centering}
\vspace*{-6mm}
\caption{\\Component 21}\label{fig:c21_wc}
\end{minipage}\qquad
\begin{minipage}[b]{.165\textwidth}
\includegraphics[width=\textwidth]{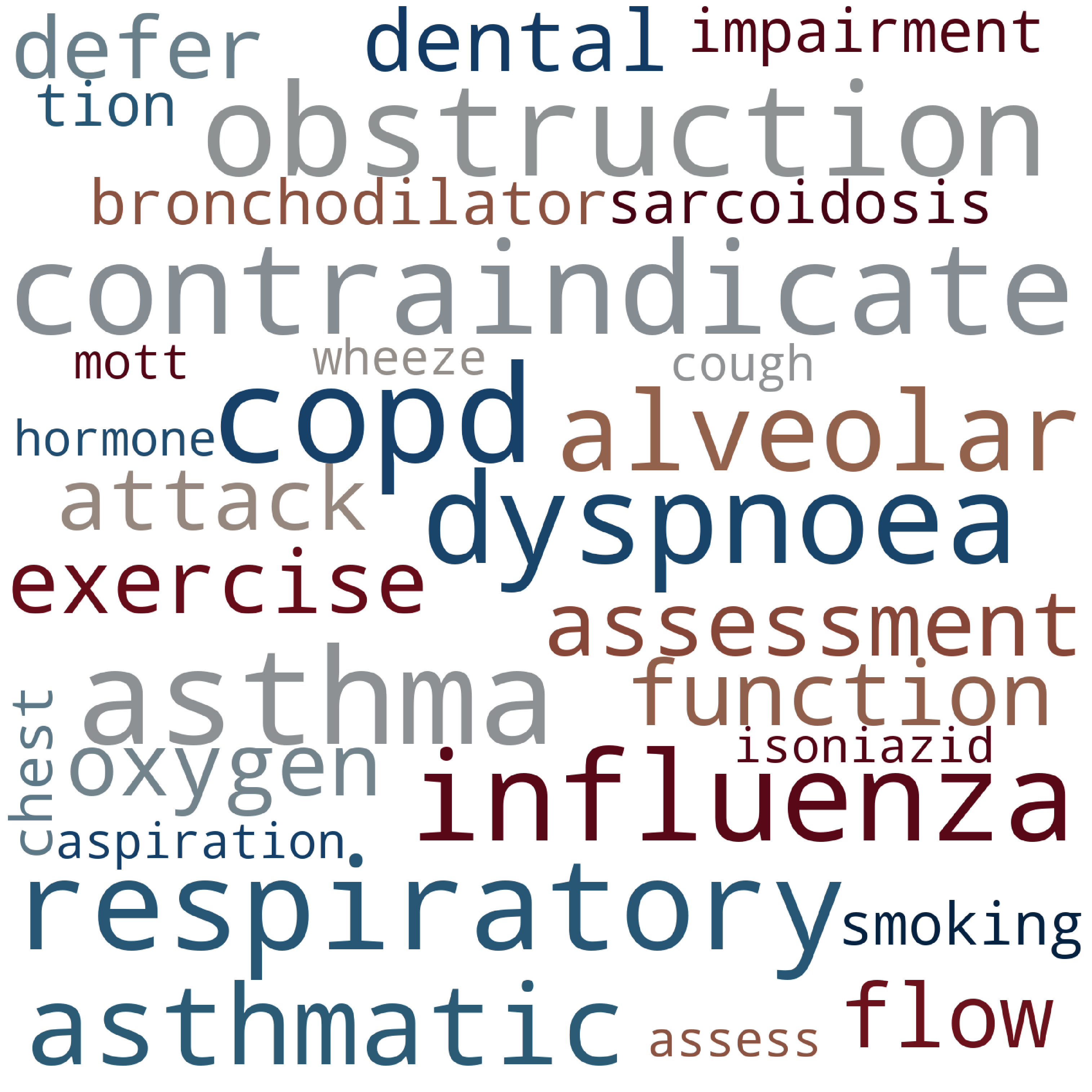}
\captionsetup{justification=centering}
\vspace*{-6mm}
\caption{\\Component 56}\label{fig:c56_wc}
\end{minipage}\qquad
\begin{minipage}[b]{.165\textwidth}
\includegraphics[width=\textwidth]{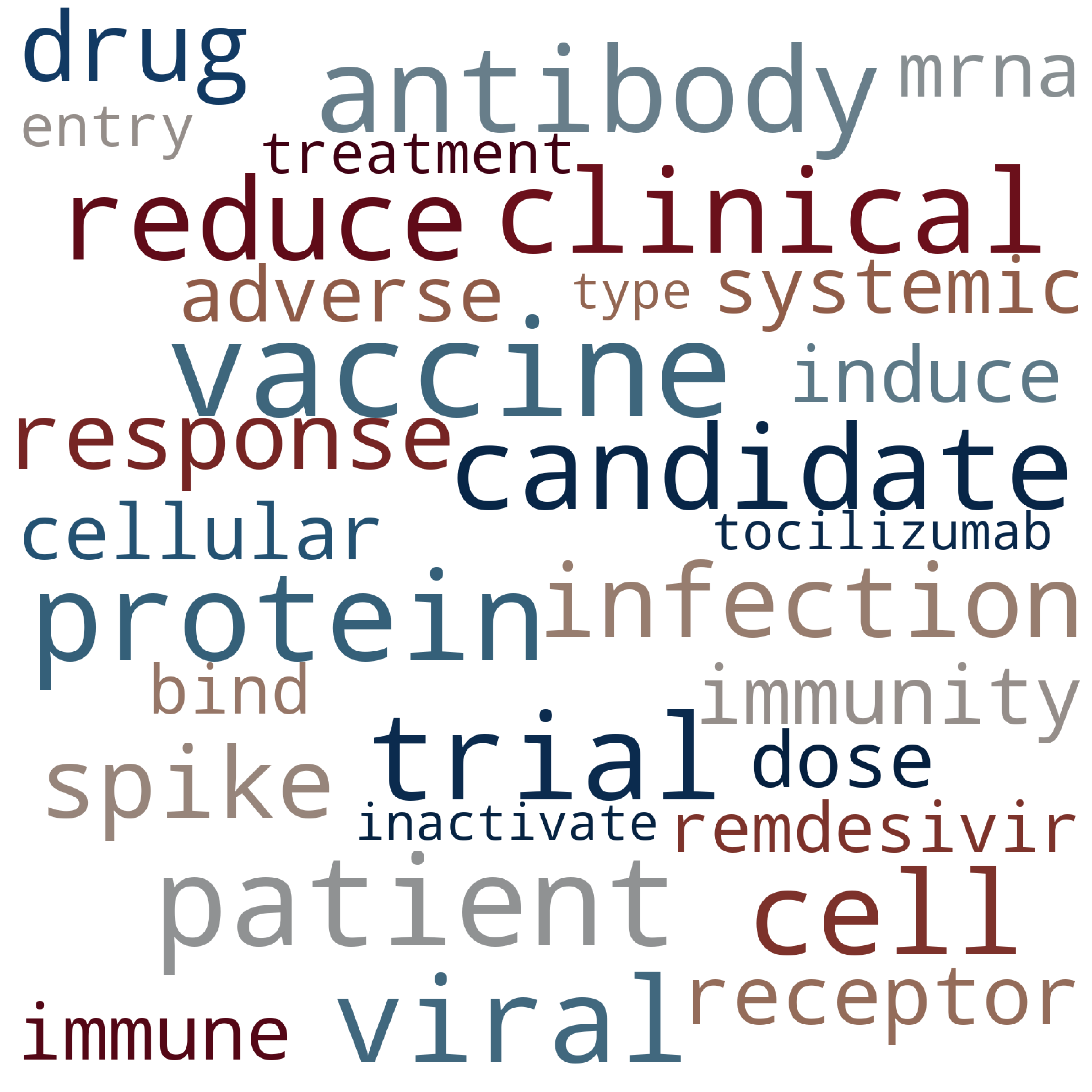}
\captionsetup{justification=centering}
\vspace*{-6mm}
\caption{\\Component 66}\label{fig:c66_wc}
\end{minipage}
\end{figure*}

\begin{figure*}[t!]
\centering
\includegraphics[width=1\textwidth]{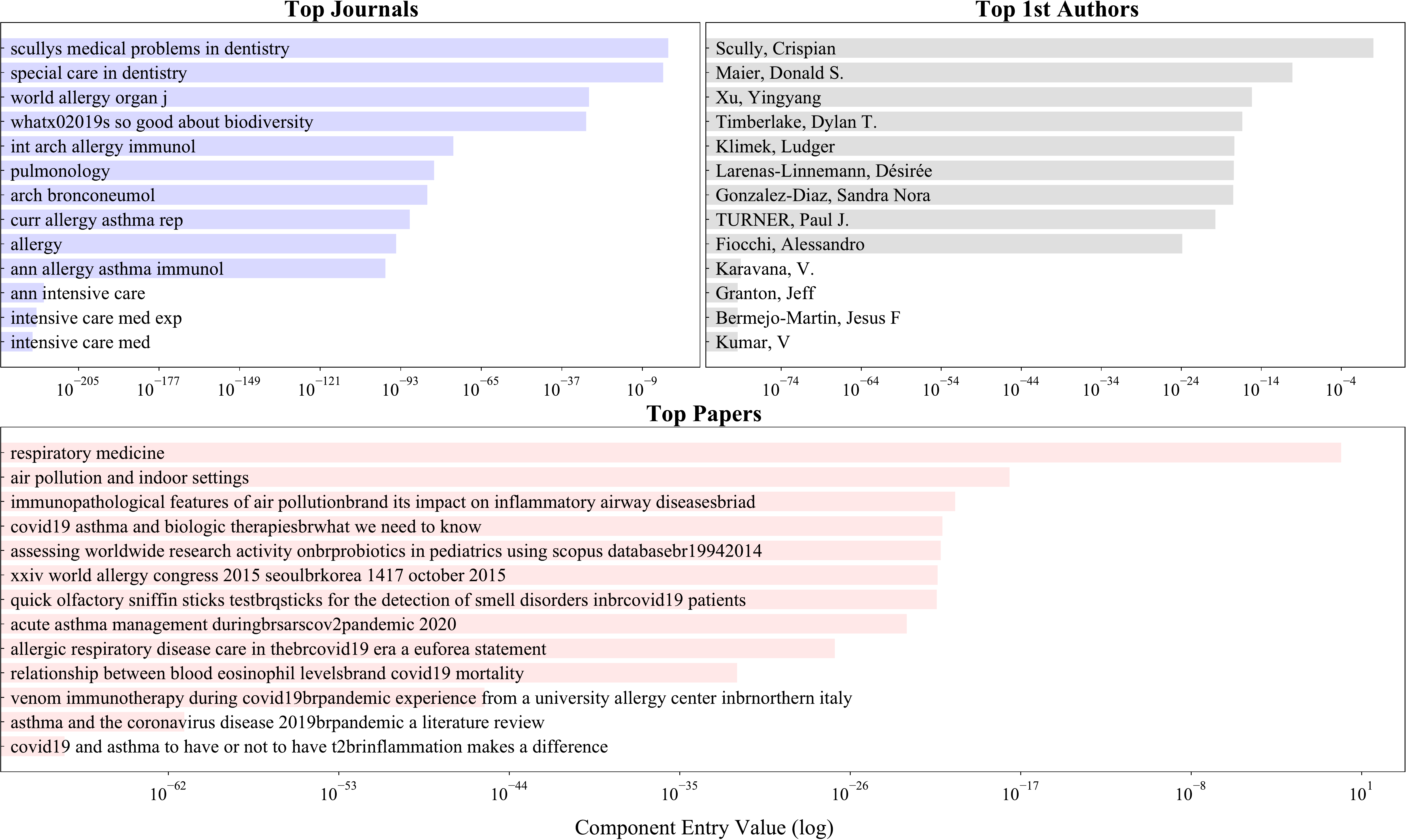} 
\vspace*{-6mm} 
\caption{The top 13 values in the latent factors of component 56. The topic keywords for this component are shown in Figure \ref{fig:c56_wc} as a word-cloud. Tensor decomposition placed the papers and journals related to air, asthma, and pollution in this component. The \textit{Journals} factor shows similar journals in close proximity. The authors in this component include the ones who have published the listed papers in same or relevant journals. \label{fig:c56_component}}
\end{figure*}


In our analysis, we use the Canonical Polyadic (CP) decomposition. For a $d$ dimensional tensor, CP decomposition is written as:

\begin{equation}
\bm{\mathscr{X}} \approx \bm{\mathscr{M}} = \llbracket \lambda \; ; \; \text{\textbf{A}}^{(1)}, \text{\textbf{A}}^{(2)},\cdots,\text{\textbf{A}}^{(d)} \rrbracket 
\end{equation}
where $\bm{\mathscr{M}}$ is the low-rank $R$ approximation of $\bm{\mathscr{X}}$, and computed as:

\begin{equation}
\bm{\mathscr{M}} \equiv \sum^{R}_{r=1} \lambda_r \text{\textbf{a}}^{(1)}_{r} \circ \text{\textbf{a}}^{(2)}_{r} \circ \cdots \circ \text{\textbf{a}}^{(d)}_{r}
\end{equation}
where $\circ$ is the outer product of the latent factors $\text{\textbf{a}}^{(d)}_{r}$ that are normalized to sum up to 1, and the weight is absorbed by each $\lambda_r$. Finally, $\textbf{A}^{(d)}$ is the set of $R$ latent factor vectors for dimension $d$:
\begin{equation}
\textbf{A}^{(d)}=\{\text{\textbf{a}}_{1}^{(d)}, \text{\textbf{a}}_{2}^{(d)}, \dots, \text{\textbf{a}}_{R}^{(d)}\}
\end{equation}

For further information on tensors and CP in particular, see \cite{KoBa09}. We build an order four tensor $\bm{\mathscr{X}}$ shaped \textit{105300} x \textit{128359} x \textit{10321} x \textit{821410}, where the dimensions of the tensor are \textit{1st Author} x \textit{Document} x \textit{Journal} x \textit{Words}. Here the \textit{Document} dimension represents the title of the articles. An entry in this tensor is $\bm{\mathscr{X}}_{a,p,j,w} = log(1+x)$, where $x$ is number of times the first author $a$ used the word $w$ in document $p$ that was published in journal $j$. There are approximately $1.15 x 10^{20}$ entries in $\bm{\mathscr{X}}$, but only 63,418,308 (or $5.53 x 10^{-11}$\%) of them are non-zero. Hence, we exploit this extreme sparsity, and store $\bm{\mathscr{X}}$ in \textit{COO} format where only the non-zero entries are represented by a list of coordinates along with the list of non-zero values for each coordinate.

We factorize this tensor with the CP Alternating Least Squares (CP-ALS) algorithm \cite{Battaglino2018APR, TTB_Software}\footnote{CP-ALS is available with MATLAB Tensor Toolbox: \url{https://www.tensortoolbox.org/cp_als_doc.html}}. $\bm{\mathscr{X}}$ is decomposed for ranks $R = i \in K: K = \{20, 40, 60, 80, 100, 120, 200\}$. Since the extracted latent factors for the \textit{Words} dimension represent the topic keywords, we employ the \textit{cosine similarity} scores when comparing each of the factor vectors $\text{\textbf{a}}_{r}^{(\text{Words})}$, where $r \in \{1, \cdots, \text{\textit{i}}\}$ and $i \in K$, to limit the components to unique topics. We use the \textit{cosine similarity} metric because it measures how much two factors point in the same "direction" (i.e. how similar their entry value distributions are).


After filtering the components with a (somewhat arbitrary) \textit{cosine similarity} score threshold of $0.35$ or higher, we select 73 out of 620 components. During our analysis, we look at each of the 73 components, and plot the entries with the highest values for three of the latent factor vectors $\{\text{\textbf{a}}_{r}^{(\text{\textit{1st Author}})}, \text{\textbf{a}}_{r}^{(\text{Document})}, \text{\textbf{a}}_{r}^{(\text{Journal})}\}$ to observe the groupings of relevant documents, journals, and authors. The latent factors for the fourth dimension, $\text{\textbf{a}}_{r}^{(\text{Words})}$, are used to form the word-clouds representing the topic keywords.


\section{Results}
\label{sec:results}

Modeling the corpus in a multidimensional space allowed us to analyze the results using the information from each dimension simultaneously. Another benefit of using tensor decomposition over the various "black-box" machine learning methods is the interpretability of the results. The values in the extracted latent factors can indicate meaningful relationships in the data. We therefore report our findings by manually inspecting the paper, journal, author, and word groupings in the components.

We first look at the word clouds of topic keywords obtained from the \textit{Words} dimension in each component. In Figures \ref{fig:c7_wc}, \ref{fig:c16_wc}, \ref{fig:c21_wc}, \ref{fig:c56_wc}, and \ref{fig:c66_wc} we see that the relevant words are grouped. Specifically, Figure \ref{fig:c7_wc} includes terms pertaining to public health. Terms concerning education during the pandemic are grouped in Figure \ref{fig:c16_wc}. Figure \ref{fig:c21_wc} contains information regarding aviation security. Terms related to vaccination are collected in Figure \ref{fig:c66_wc}. Indeed, the extracted topic keywords semantically parallel the documents, journals, and author publications grouped in each of the respective components. We provide an example of this with Component 56 in Figure \ref{fig:c56_component} which focuses on respiratory studies as shown in Figure \ref{fig:c56_wc}.

Component 56 is one of our more interesting results. As the keywords indicate, the papers and journals listed in this component focus on respiratory issues caused and/or exacerbated by COVID-19. Using our interactive visualization, we identify patterns preserved by the tensor factorization. Journals, papers, and author publications in this component address topics such as "asthma", "air pollution", and "allergy". While the majority of the papers belong to a single journal, "World Allergy Organization Journal", we were also able to cluster other relevant journals such as "International Archives of Allergy and Immunology" and "Current Allergy and Asthma Reports". We also notice distinct groupings of journals concerning other research fields in the same component. 

Several components grouped together single authors who specialize in a niche area of research. For instance, one component grouped together a single author "Ruwantissa Abeyratne" who has produced several articles on aviation law. Furthermore, CORD-19 includes textbooks divided into individual chapters and our methodology produced several components that identified and grouped together textbooks by the same author. More prolific authors (most notably "Theodore Tulchinsky") have multiple components dedicated to their textbooks. Despite this, we were able to extract relevant keywords for these components. For instance, the word cloud for Tulchinsky's components includes words pertaining to public health policy.

\section{Conclusion}
In this paper, we expanded upon our prior work in organizing the COVID-19 literature using tensor analysis. We showed that using a higher-order representation of the documents allow capturing of the topic keywords for the papers found in the CORD-19 corpus. We also observed groupings of similar articles, journals, and researchers within and among the components obtained by taking the CP decomposition of our four dimensional CORD-19 tensor. Future work can consider using a non-negative tensor factorization algorithm instead of CP-ALS to improve interpretability, and explore different combinations of tensor dimensions.

\section{Acknowledgment}
This manuscript has been approved for unlimited release and has been assigned LA-UR-21-25094. This research was partially funded by the Los Alamos National Laboratory (LANL) Laboratory Directed Research and Development (LDRD) grant 20190020DR and LANL Institutional Computing Program, supported by the U.S. Department of Energy National Nuclear Security Administration under Contract No. 89233218CNA000001.

\bibliographystyle{ACM-Reference-Format}
\bibliography{references.bib}


\end{document}